# Handling Heavily Abbreviated Manuscripts: HTR engines vs text normalisation approaches

Jean-Baptiste Camps(✉)[0000−0003−0385−7037], Chahan Vidal-Gorène[0000−0003−1567−6508], and Marguerite Vernet

École nationale des chartes – Université Paris, Sciences & Lettres, 65 rue de Richelieu 75002 Paris, France
{firstname.lastname}@chartes.psl.eu
http://www.chartes.psl.eu/

**Abstract.** Although abbreviations are fairly common in handwritten sources, particularly in medieval and modern Western manuscripts, previous research dealing with computational approaches to their expansion is scarce. Yet abbreviations present particular challenges to computational approaches such as handwritten text recognition and natural language processing tasks. Often, pre-processing ultimately aims to lead from a digitised image of the source to a normalised text, which includes expansion of the abbreviations. We explore different setups to obtain such a normalised text, either directly, by training HTR engines on normalised (i.e., expanded, disabbreviated) text, or by decomposing the process into discrete steps, each making use of specialist models for recognition, word segmentation and normalisation. The case studies considered here are drawn from the medieval Latin tradition.

## 1 Introduction

### 1.1 Abbreviations in Western Medieval Manuscripts

In medieval Latin manuscripts, abbreviations are fairly common and follow a practice that was established, by and large, during the first centuries A.D., reserved for a time to administrative and everyday written production and then extended to literary manuscripts [11,10]. They mostly derive from two antique conventional systems: *notae antiquae*, on the one hand, that proceed by suspension, superscript letters or tachygraphic signs, and mostly affect grammatical morphemes such as inflections, adverbs, prepositions, pronouns and the forms of the verb *esse*; Christian *nomina sacra*, on the other, abbreviations of holy names, by contraction [1]. Extended by Irish monks, with the addition of new (insular) signs such as ÷ (*est*) and then standardised and generalised by the Carolingian *renovatio*, this system forms the basis for abbreviation practices in medieval Latin manuscripts, but also for the abbreviations of many vernaculars. They were notably adapted to Old French by Anglo-Norman scribes [10]. Inherited from this history are abbreviations that can be categorised as **tachygraphic sign** e.g. Tironian ⁊ (*et*) or ꝯ (*cum*, *con-*, *com-*,...).
**superscript letter** e.g. superscript *a* for *ua* or *ra* in q̊ (*qua*), t̊ns (*trans*).
**suspension** e.g. ẽ (*est*).
**contraction** e.g. D̃S (*Deus*).

In the 12[th] and 13[th] centuries, the intellectual flourishing and the development of schools and universities caused a heavy demand for written artefacts. The copying of manuscripts expanded beyond the sole framework of monastic *scriptoria* and spread to the city in professional workshops and lay *scriptoria*. The development of a larger literate milieu of students and masters, and the growth of book production led to modifications in intellectual practices and ultimately in the processes of reading and writing. A switch, at least in scholastic milieus, from slow syllabic reading to faster, expert modes of reading, based on the global perception of each word, led to a very significant increase in word-level abbreviations, and, specifically, abbreviations by contraction [10]. They display much variety and include:

**simple contraction** using letters from the original word (with a marker to make the presence of the abbreviation explicit) that can be relatively **unambiguous**, e.g. eccl̃a (*ecclesia*), r̄oe (*ratione*) or **ambiguous**, e.g. ĩ for *ita*, *illa* or *infra*, or even sometimes *prima* or *una*, depending on the context.
**composite contraction** combining other conventional devices with the contraction itself, for instance p̊ (*persona*).

In Latin manuscripts, we already encounter a great versatility of signs and many homographic abbreviations (or alternative expansions), a situation that is made even more uncertain in vernacular manuscripts, due to spelling variation.

### 1.2   Expanding abbreviations

Expanding abbreviations is not a trivial task because there is no unambiguous character-, syllable- or word-level mapping between abbreviation and expansion: the same abbreviation can correspond to several expanded forms, and an expanded form can have several abbreviations. In addition, the same sign can fulfil alternative functions on different levels. Attempts to model the relationship between abbreviations and expanded forms exist on a theoretical level in linguistic research [16], but the situation remains complex. In graph theory terms, the binary relation between the set of abbreviations, and the set of expanded forms can be characterised as a many-to-many relation and not as a function. On an individual level, a sign can have

**one character expansion** e.g., ⁊ → *et* (word) or *-et* (word syllable) in Latin; while on the contrary ⁊ → *et*, *e*, *ed* (Old French).
**multiple character expansion** ꝯ → *cum* (word) and *cum-*, *con-*, *com-* (prefix) in both Latin and Old French.

The same is also true at the level of the sequence of signs, that can have **one expanded form** e.g., r̃õe → *ratione* (Latin).
**many expanded forms** i̊ → *ita, illa, infra, prima, una...* (Latin).

The same sign can enter into different relations both in isolation and as part of groups. A good example of this plasticity is given by the common abbreviative mark known as 'titulus' or 'tittle': alone, it can be used to stand for a nasal consonant (-*m*- or -*n*-), while it is also the most common marker to indicate that a word is globally abbreviated, having, in that case *no explicit character value* per itself, for instance in the aforementioned r̃õe example. The actual incidence of abbreviation polyvalence varies in time, between languages and language variants, as well as per types of documents or texts, and ultimately, scribes.

### 1.3 Computational approaches

Handling abbreviations is a general problem with manuscripts, especially medieval manuscripts, but we find relatively few studies dealing with computational approaches to their expansion. Romero et al. report on recording both diplomatic transcriptions (with abbreviations) and normalised (expanded) transcriptions of Dutch Medieval manuscripts, through the use of XML/TEI, but give only results for the first version [19].

The problem of homograph abbreviations and the versatility of signs seems to call for a representation of context. In practice, two main kinds of approaches have been used: HTR systems trained on normalised data on one hand; treating abbreviation expansion as a text normalisation task on the other.

HTR approaches can include some representation of token context, because state-of-the-art HTR systems usually take into account the full text line. From a pragmatic perspective, this makes the creation of ground truth easier, because it facilitates the reuse of existing transcriptions and has been investigated for this very reason, yet tended to show relatively high character error rate (CER), where 'deletions' (including letters that should have been added as part of abbreviation expansion) represent more than half of the errors [25,2].

Alternatively, normalisation can be treated as a separate (posterior) normalisation task, based on the output of the HTR phase. The literature concerning historical text normalisation is considerably larger, and includes approaches based on substitution lists, rules, as well as distance-based, statistical approaches (in particular, character-based neural machine translation) and more recently neural models [4]. To include a modelling of context, normalisation systems can reuse deep-learning architectures originally intended for neural machine translation [4,9] or lemmatisation [6,15].

In this paper, we plan to explore two approaches to expand abbreviations in Latin manuscripts, with and without post-processing. Evaluations are carried out with a small dataset in order to highlight benefits of each approach within the scope of an under-resourced language.

## 2   MS BnF lat. 14525

For this work, we used MS BnF lat. 14525, the subject of an ongoing master's thesis [22]. It belonged to the library of Saint-Victor of Paris, a canonical Abbey that played a central role in the intellectual life of Paris, especially during the 12$^{th}$ century, and was situated at the intersection between the monastic and Parisian worlds.

BnF lat. 14525 was produced for the library of Saint-Victor in the first half of the 13$^{th}$ century. It is not made up of a single codicological unit, but was completed and improved about ten years later. This manuscript of 305 folios brings together various texts from different origins (Victorian, Cistercian, Parisian schools) and includes spiritual treatises, works on practical theology, numerous sermons, Constitutions of the Fourth Lateran Council or synodal constitutions, and even particular material and spiritual privileges. Despite this heterogeneity, it is a very useful item for a better understanding of Saint-Victor during this time. About ten different hands wrote it, though the handwriting is quite similar. We are dealing here with a script akin to scholastic writing, with a few broken stems, a reduced module, as well as numerous abbreviations.

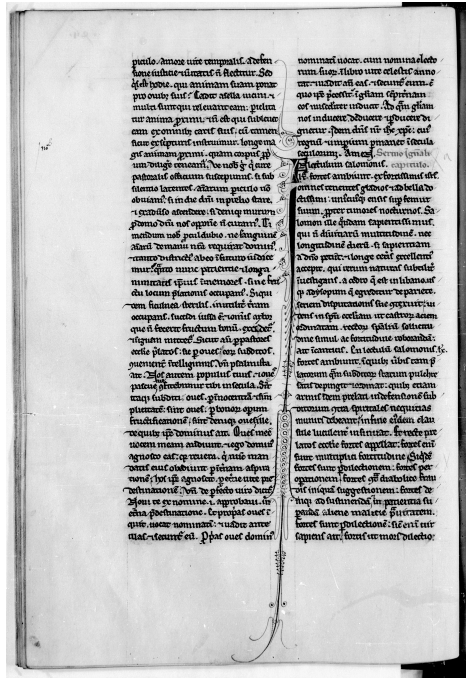

**Fig. 1.** Folio 45v, MS BnF lat. 14525

We focus here on one of these hands, which occupies a quarter of the manuscript, i. e. 79 fols, copied on two columns of 42 ruled lines (the first of which also carries writing), most likely between 1215 and 1225, on parchment of relatively good quality despite some defects. The copy is neat with few errors, and corrections by expunctuation or crossing out, as well as some interlinear or marginal additions. Hyphens and dotted 'i's are also present, although irregularly. The writing itself is very regular and skilled, and the use of abbreviations is quite essential. The density of abbreviations, measured as the ratio of words with at least one abbreviation, is around 47%, not out of the ordinary for this type of manuscript, but much larger than what is found for instance for contemporary Old French epic manuscripts, with figures in the 10 to 20% range [5]. We find the systematic abbreviation of *et* in two forms: & and ꝫ. The copyist uses different tittles:

mostly *er*, while ¯ stands for nasals or signals suspensions or contractions. ˘ is used for *ur* and ˀ for *us*. Superscript letters are also abundantly used, in particular *a*, *o* and *i*. This frequent use of abbreviations is typical of the kind of scholastic writing whose use was widespread from the mid-13th century in university circles.

Images of the manuscript are available on Gallica as a grayscale digitisation of the microfilm [3].

## 3 Experiments and results

The paper aims to compare two approaches to decipher Latin abbreviations, and HTR experiments have been carried out with two neural architectures. The first one (HTR-CB) is proposed by Kraken [13].The results produced by this architecture serve as a baseline for pure character-level recognition of this manuscript and are provided to other modules in the defined pipeline (see *infra* 3.3). The second architecture (HTR-WB) is an adaptation of that proposed on Calfa Vision [24], originally developed for the reading of medieval Armenian manuscripts and the management of abbreviations and ideograms specific to this language and which cannot be recognised at the character level [23]. Character recognition is preceded by a word-based system, to which we first provide an exhaustive list of abbreviations encountered in the manuscript. The results of this architecture serve as a point of comparison.

### 3.1 Ground truth creation

Training and testing data have been annotated with a layout analysis and a baseline model, and manually proofread via eScriptorium [14,12]. The dataset is composed of 1.861 lines of text (1,524 reserved for training, 168 for validation and 169 for testing). We built a total of four datasets:

1. **D-exp**, consisting of a transcription with full expansion of the abbreviations. We consider two variants, **D-exp1** with inter-word spaces restored (separation of words according to Latin grammar and not the spacing present in the manuscript), and **D-exp2** without spaces. These datasets are respectively composed of 34 and 33 classes. The number of classes has been limited to include enough samples in each. This can lead to strong intra-class inertia, due in particular to the unsystematic grouping of upper- and lower-case letters for the less endowed classes.
2. **D-abb**, composed of transcriptions registering the abbreviation system used in this manuscript (see *supra* 1.1). We also consider two variants, **D-abb1** with inter-word spaces, and **D-abb2** without spaces. These datasets are respectively composed of 60 and 59 classes. In detail, we have 36 classes representing alphanumeric characters and punctuation marks, 10 classes specific to Latin paleography to represent certain abbreviations according to the same scheme as the Oriflamms project (e.g. ƀ and ꝗ; see *infra*), and 13 classes made up of combining signs (written above or below one or more letters). These last 13 classes can be difficult to identify for an HTR engine.

Moreover, additional data was used for some of the trainings:

1. **Oriflamms** diplomatic and allographetic transcriptions provided by D. Stutzmann and team [21,7,18,20].
2. **PL** 216 volumes of normalised editions from Migne's *Patrologia latina* [17].

### 3.2  HTR on abbreviated and expanded data

Common parameters have been chosen for training steps of HTR-CB and HTR-WB. Line images provided as input are resized to 64px in height and have varying widths. No Unicode normalisation or data augmentation is applied. A repolygonisation has been performed to equalise polygons of first and last rows of each text columns [24]. Due to the small size of the dataset and to avoid overfitting, training steps are carried out with a dynamic learning rate starting at 0.001 – to which a coefficient of 0.75 is applied every 10 epochs –, and we use a batch of one image for each iteration. We have limited training to 30 epochs. First results are summarised in Table 1.

**Table 1.** Evaluation of character-based and word-based HTR system on datasets with and without expanded abbreviations.

|         | CER (%) | | | |
|---------|---------|---------|---------|---------|
|         | D-exp1  | D-exp2  | D-abb1  | D-abb2  |
| HTR-CB  | 10.59   | 9.69    | 4.89    | 4.55    |
| HTR-WB  | 3.76    | 2.96    | 5.57    | 4.83    |

The two architectures give equivalent results on the two variants of the D-abb dataset. At identical initial parameters, we do not observe any significant difference, except that HTR-CB converges twice as fast as HTR-WB. A dynamic learning rate brings a real benefit for both of two architectures (stagnation of the CER until epoch 10 then gradual reduction). Epoch 30 is never reached. 20% of the errors of HTR-CB are focused on combining signs, but it generally recognises combined letters well. It also mistakes close classes as p̣ and p, or q̣ and q, but the lack of data can be a good explanation of this phenomenon.

There is a clear benefit to using a word-based approach for the management of abbreviations directly within the HTR process. If it seems quite logical that the absence of spaces, an ambiguous notion in manuscripts, can really benefit to text recognition, there is however only a marginal gain between D-abb1 and D-abb2. On the other hand, the HTR-WB model takes advantage of the lack of spaces. Most errors are focused on small and independent abbreviations, generally limited to one single character (e.g. m̊ > modo), but also on long abbreviated words (e.g. micd̃ia > misericordia) for which we do not have enough samples in the training set (e.g. only one sample of misericordia). We give in Table 2 an example of predictions. For the rest of the paper, we consider the HTR-CB output, as shown in Table 2.

**Table 2.** Example of predictions with and without abbreviations.

| | |
|---|---|
| | ![manuscript line] |
| GT-exp1 | pro domo domini nos opponere non curamus. ti |
| **HTR-WB** | prodomo domini nos oponere non curamus. ti |
| GT-abb2 | ꝑdomodñinosoppoñeñcuram̊.ti |
| **HTR-CB** | ꝑdomodñinosoppoñeñcuram̊.tiē |

### 3.3 Text normalisation approach

The modular text normalisation approach uses several consecutive steps, with specialised tools each necessitating a training of its own [6], as follows (fig. 2):

1. **HTR** (*see above*).
2. **word segmentation** using a deep-learning word segmentor, `Boudams`. [8].
3. **Abbreviation expansion and word normalisation** using a deep learning based word annotator, `Pie`. [15].

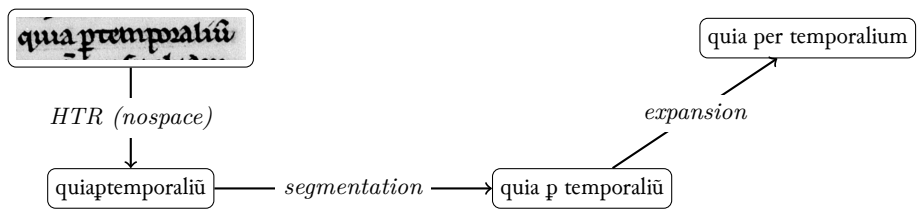

**Fig. 2.** Workflow for the full modular approach [6]. The digital image is segmented in lines that are submitted to recognition, and the recognised text then goes through word segmentation and normalisation (abbreviation expansion).

*Word segmentation* A neural word segmentor, `Boudams`, was trained using the best known configuration: convolutional encoder without position embeddings (Embedding 256, KernelSize 5, Dropout 0.25, Layers 10) [8]. Three datasets were used: for abbreviated text, D-abb (with word separation and line hyphenations normalised) alone, and with the addition of Oriflamms data; for normalised version, D-exp (with word separation and line endings normalised) with the addition of PL data. Three models were trained for each setting, and the best of the three (based on F-statistic) was selected. Results are shown in Table 3.

**Table 3.** Results for training of the word segmentor `Boudams`

| | | n-lines | | | test scores | | |
|---|---|---|---|---|---|---|---|
| dataset | type | train | dev | test | F-s. | Prec. | Recl. |
| D-abb | abbr. | 1 962 | 196 | 200 | 95.1 | 96.1 | 94.1 |
| D-abb+Oriflamms | abbr. | 17 139 | 196 | 200 | 97.3 | 97.4 | 97.1 |
| D-exp+PL | norm. | 3 891 929 | 196 | 200 | 98.8 | 98.6 | 98.9 |

*Word normalisation* For abbreviation expansion, the neural tagger Pie was trained [15], following a setup already used for Old French [6]. It was trained on an aligned version of the previous D-abb and D-exp (with word separation and line hyphenations normalised) alone, and with the addition of the Oriflamms data. Results are shown in Table 4.

**Table 4.** Results for the training of Pie for abbreviation expansion

| dataset | Test accuracy (%) |||||
|---|---|---|---|---|---|
|  | all | known | unkn. | ambig. | unkn. targ. |
| **D-abb/D-exp** | 94.04 | 95.37 | 92.42 | 89.58 | 91.23 |
| **D-abb/D-exp+Oriflamms** | 97.02 | 98.65 | 95.02 | 97.92 | 92.86 |

### 3.4   Results

The full pipeline was evaluated globally on the normalised transcription of one folio of the manuscript, with expanded abbreviations, normalised word segmentation and line hyphenations. The metrics used were character error rate (CER) and word error rate (WER) in the Python `fastwer` implementation [26]. Results are shown in Table 5.

**Table 5.** Results of the application of the full pipeline for a selection of the most accurate setups, with or without additional data.

| | Setups |||||| Scores ||
|---|---|---|---|---|---|---|---|---|
| | **HTR** || **Segment.** || **Norm.** || **Test** ||
| | Data | Soft. | Data | Soft. | Data | Soft | CER | WER |
| 1a | D-exp1 | CB (HTR-CB) | | | | | 15.63 | 82.38 |
| 1b | D-exp1 | WB (HTR-WB) | | | | | **7.46** | **54.68** |
| 2a | D-exp2 | CB (HTR-CB) | D-exp1+PL | Boudams | | | 12.65 | 50.83 |
| 2b | D-exp2 | WB (HTR-WB) | D-exp1+PL | Boudams | | | **6.89** | **34.01** |
| 3a | D-abb2 | CB (HTR-CB) | D-abb1 | Boudams | D-abb/exp | Pie | 10.89 | 41.28 |
| 3b | D-abb2 | CB (HTR-CB) | +Orifl. | Boudams | +Orifl. | Pie | **8.60** | **31.81** |

The simple use of an HTR engine, trained on a normalised transcription with normalised spaces and expanded abbreviations actually provides a strong baseline. This is particularly true of the word-based HTR, and in terms of character error rate. Yet, they are overperformed by the more refined setups, either in terms of character error rate or word error rate. The best scores (Table 5) are obtained, for the character error rate, by the HTR-WB trained on normalised transcriptions whose output is then resegmented by a word segmentor trained on the *Patrologia latina*; for the word error rate, by the HTR-CB trained on abbreviated transcriptions, whose output is then resegmented and normalised with models trained using the additional Oriflamms data.

**Table 6.** Facsimile and ground truth with sample predictions of the three best performing setups (word with content or segmentation errors **in bold**), from the beginning of fol. 45v (see fig. 1).

| Facsimile | GT |
|---|---|
| 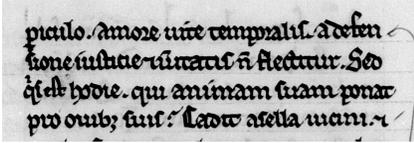 | p*eri*culo . amore uite temporalis . a defensione iusticie *et* u*eri*tatis n*on* flectitur . sed q*uis* est hodie . qui animam suam ponat pro ouibus suis : cadit asella uicini . et |

| 1b | 2b |
|---|---|
| **piculo**. amore uite temporalis. **adefensione** iusticie et ueritatis non flectitur. **sedquis** est hodie. qui animam suam **ponatpro** ouibus suis : cadit asella uicini. et | **piculo** . amore uite **temprus** . a **de sensione** iusticie et ueritatis non flectitur . sed quis est hodie . qui animam suam ponat pro ouibus suis : cadit asella uicini . et |

| 3b | |
|---|---|
| **pculo** . **amoreure temporauis** ; a **defsasione** iusticie et ueritatis non flectitur . **sedquas st** hodie . qui animam suam ponat pro ouibus suis : **cadita sella** uicini . et | |

## 4 Discussion and further research

Results seem to confirm the importance of the word level rather than the character level, with word-based overperforming character-based HTR, and with character-based HTR results being improved by adding tools dealing with word-level tokens and context, in particular for a small dataset. In this regard, artificial intelligence can be compared to human intelligence and seems to confirm the practice of global reading and global perception of the words rather than individual letters, reflected in the use of abbreviations in Latin (especially scholastic) manuscripts. Future research should investigate differences with vernacular manuscripts, for instance literature in Old French, where reading and the use of abbreviations are supposed to have remained mostly syllabic.

Another conclusion of this paper is that, given the nature of spacing in medieval manuscripts, the word error rate regarding normalised words can be greatly reduced by using a dedicated word segmentor, for which (normalised) training data can be easily collected (since all that is needed are normalised editions). In our experiments, this proved to be the best performing setup in terms of character error-rate.

Thirdly, a fully modular approach combining HTR on abbreviated data with a word segmentor and a text normalisation tool is the best performing in terms of word error rate. It would be possible to improve further the results of this approach by improving first the word-based approach – that still suffer of a lack of data to manage with the word level –, and then the segmentation and normalisation steps with additional data, yet the limit to this approach is the availability of adequate training material, i.e., editions recording abbreviations, that are much harder to come by than normalised editions.

Another future line of research should pursue the comparison with the vernacular, where the ambivalence of abbreviations, due to the variety of written norms and alternative spellings, should make the automated production of a normalised text more difficult. It should also investigate the impact of data augmentation techniques, particularly easy for the word segmentation training that could, for instance, include random character substitutions as to emulate HTR effect [8].

**Acknowledgements** We thank the École nationale des chartes and the DIM STCN for the computing power and GPU server used for training, as well as INRIA and Calfa. We also thank Marc H. Smith for his keen review of our draft. Any remaining mistakes are only attributable to us.

**Datasets availability** Datasets produced for this paper and evaluation scripts are available at DOI 10.5281/zenodo.5071963.